\documentclass{article}

\usepackage[T1]{fontenc}
\usepackage{microtype}
\usepackage{graphicx}
\usepackage[export]{adjustbox}
\usepackage{xcolor}
\usepackage{amsmath}
\usepackage{subfigure}
\usepackage{tablefootnote}
\usepackage[inline]{enumitem}
\usepackage{booktabs} 
\usepackage{multirow}
\usepackage{hyperref}

\usepackage[accepted]{icml2020}

\usepackage{amsmath,amsfonts,amssymb,amsthm,dsfont}
\usepackage{tikz}
\newcommand*\circled[1]{\tikz[baseline=(char.base)]{
            \node[shape=circle,draw,inner sep=1pt] (char) {#1};}}

\def\R{\mathbb{R}}

\def\1{\mathds{1}}

\def\A{\mathcal{A}}
\def\CA{\mathcal{CA}}
\def\L{\mathcal{L}}

\def\X{\mathcal{X}}
\def\Z{\mathcal{Z}}
\def\E{\mathcal{E}}

\def\D{\mathcal{D}}
\def\L{\mathcal{L}}

\def\our{LoCondA}

\newcommand{\gray}[1]{\textcolor{gray}{#1}}

\icmltitlerunning{Modeling 3D Surface Manifolds with a Locally Conditioned Atlas}

\ifdefined\isaccepted
    \newcommand{\code}{\url{https://github.com/gmum/LoCondA}}
\else
    \newcommand{\code}{<hidden during review>}
\fi

\begin{document}

\twocolumn[
\icmltitle{Modeling 3D Surface Manifolds with a Locally Conditioned Atlas}



\icmlsetsymbol{equal}{*}

\begin{icmlauthorlist}
\icmlauthor{Przemysław Spurek}{to,ide}
\icmlauthor{Sebastian Winczowski}{to}
\icmlauthor{Maciej Zięba}{boo}
\icmlauthor{Tomasz Trzciński}{goo,ide}
\icmlauthor{Kacper Kania}{goo}
\icmlauthor{Marcin Mazur}{to}
\end{icmlauthorlist}

\icmlaffiliation{to}{Faculty of Mathematics and Computer Science, Jagiellonian University, Kraków, Poland}
\icmlaffiliation{goo}{Warsaw University of Technology, Warsaw, Poland}
\icmlaffiliation{boo}{Wroclaw University of Science and Technology}
\icmlaffiliation{ide}{IDEAS NCBR}

\icmlcorrespondingauthor{Przemysław Spurek}{przemyslaw.spurek@uj.edu.pl}
\icmlkeywords{Machine Learning, ICML}

\vskip 0.3in
]


\printAffiliationsAndNotice{}  

\begin{abstract}

Recently proposed 3D object reconstruction methods represent a mesh with an \textit{atlas} - a set of planar patches approximating the surface. However, their application in a real-world scenario is limited since the surfaces of reconstructed objects contain discontinuities, which degrades the quality of the final mesh. This is mainly caused by independent processing of individual patches, and in this work, we postulate to mitigate this limitation by preserving local consistency around patch vertices. To that end, we introduce a Locally Conditioned Atlas (\our{}), a framework for representing a 3D object hierarchically in a generative model. Firstly, the model maps a point cloud of an object into a sphere. Secondly, by leveraging a spherical prior, we enforce the mapping to be locally consistent on the sphere and on the target object. This way, we can sample a mesh quad on that sphere and project it back onto the object’s manifold. With \our{}, we can produce topologically diverse objects while maintaining quads to be stitched together. We show that the proposed approach provides structurally coherent reconstructions while producing meshes of quality comparable to the competitors.\footnote{We publish code of \our{} at: \code{}} 
\end{abstract}

\section{Introduction}

Efficient 3D object representations are fundamental building blocks of many  computer vision and machine learning applications, ranging from robotic manipulation~\cite{kehoe2015survey} to autonomous driving~\cite{yang2018pixor}. Contemporary 3D registration devices, such as LIDARs and depth cameras, generate these representations in the form of unordered sets of 3D points sampled sparsely on object surfaces, called \textit{point clouds}. Although a single point cloud~\cite{qi2017pointnet,qi2017pointnet++} can be used to regenerate an object's surface details~\cite{fan2017point}, it does not contain enough information about 3D points' neighborhood structure to successfully reconstruct a smooth, high-fidelity manifold of the entire surface of an object. This shortcoming limits point clouds' applicability since surface reconstructions provide an intuitive and efficient object representation, comprehensible for both humans and machines.

Recently proposed object representations address this pitfall of point clouds by modeling object surfaces with polygonal meshes~\cite{wang2018pixel2mesh,groueix2018papier,yang2018foldingnet,spurek2020hypernetwork,spurek2020hyperflow}.  They define a mesh as a set of vertices that are joined with edges in triangles. These triangles create the surface of an object. The resulting representation is efficient and easy-to-render, while at the same time it offers additional benefits, {\it e.g.} the possibility of sampling the surface at the desired resolution, and straightforward texturing in any 3D computer graphics software. To obtain such a representation, state-of-the-art approaches leverage deep learning models based on the autoencoder architecture~\cite{wang2018pixel2mesh,spurek2020hypernetwork,spurek2020hyperflow} or based on an ensemble of parametric mappings from 2D rectangular patches to 3D primitives, often referred to as an {\it atlas}~\cite{groueix2018papier,yang2018foldingnet,bednarik2020shape,deng2020better}. The former methods are limited by the topology of the autoencoder latent space distribution, {\it e.g.}, they cannot model complex structures with a nonspherical topology~\cite{spurek2020hypernetwork,spurek2020hyperflow,wang2018pixel2mesh}. Atlas-based approaches, on the other hand, are much more flexible and enable modeling virtually any surface. However, since individual mappings' consistency is not guaranteed, those methods often yield discontinuities of the reconstructed shapes and their deformation. 

\begin{figure*}
\begin{center}
\includegraphics[width=0.85\linewidth, trim=10 30 10 20, clip]{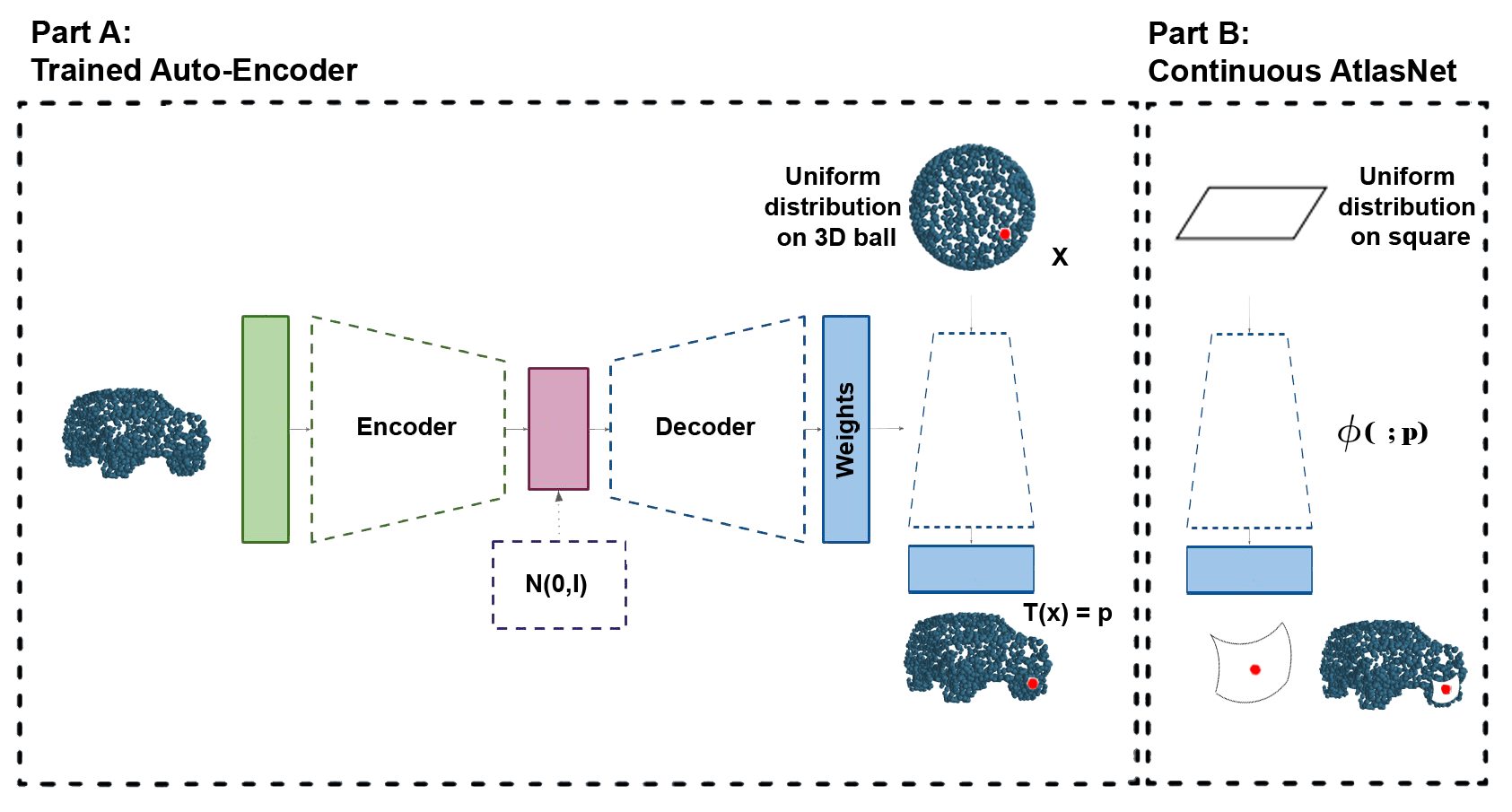}
\end{center} 
\caption{
\our{} extends a base generative hypermodel~{\bf(Part~A)} by taking a point $p$ on the surface $S$ and mapping it to a patch covering a neighborhood of $p$ (\textbf{Part~B}). 
}
\vspace{-0.5cm}
\label{fig:idea}
\end{figure*}

Although modifications proposed by \citet{bednarik2020shape} and \citet{deng2020better} improve the quality of results, their objective is to fix deformations caused by the stitching of individual mappings. We postulate that by enforcing the local consistency of patch vertices within the objective function of a model, we can avoid creating these deformations in the first place.
To that end, we propose a novel framework, \our{}, capable of generating and reconstructing high-quality 3D meshes. This framework extends the existing base hypermodels~\cite{spurek2020hypernetwork,spurek2020hyperflow} with an additional module designed for mesh generation that relies on a parametrization of local surfaces, as shown in Fig.~\ref{fig:idea}. 
Our formulation does not require the framework to possess any information about connections between points. Therefore \our{} uses only the base's data model during training, which increases the efficiency and applicability of our approach.

Practically speaking, our approach transforms the embedding of point cloud obtained from the base model to parametrize the bijective function represented by the MLP network. This function aims to find a mapping between a canonical 2D patch to the 3D patch on the surface of the target mesh. We condition the positioning and shape of a 3D patch using a single point from a point cloud generated by a base model. We repeat the procedure for each of the generated points, preserving local neighborhoods between the point cloud and the points located in the generated mesh. Intuitively, this allows us to include the stitching and reshaping of the patch within our framework's training objective, hence mitigating the possibility of shape discontinuities.   

We summarize our contributions as follows:
\begin{itemize}
    \item We propose a general framework for patch-based reconstruction methods that creates high-fidelity meshes from raw point clouds.
    \item We present \textit{Continuous Atlas} - a novel paradigm that generalizes the existing atlas methods and enables sampling any number of patches to cover any part of the reconstructed object adaptively.
    \item Finally, we show a simple conditioning mechanism for atlas-based methods that shares information between individual patches and nullifies the problem of self-intersections and holes in reconstructed meshes.
\end{itemize}

\section{Related Works}

\paragraph{3D Shape Representations}

In literature, there exist a huge variety of 3D shape reconstruction models. The most popular ones are dense, pixel-wise depth maps, or normal maps \cite{eigen2014depth,bansal2016marr,bednarik2018learning,tsoli2019patch,zeng2019deep}, point clouds \cite{fan2017point,qi2017pointnet++,yang2018foldingnet}, meshes \cite{wang2018pixel2mesh,gundogdu2019garnet,yao2020front2back,yifan2020neural}, implicit functions \cite{chen2019learning,mescheder2019occupancy,park2019deepsdf,xu2019disn,atzmon2020sal}, voxels \cite{choy20163d,hane2017hierarchical}, shape primitives \cite{chen2020bsp,deng2020cvxnet,smirnov2020deep,paschalidou2020learning}, parametric mappings \cite{yang2018foldingnet,groueix2018papier,williams2019deep,deprelle2019learning,bednarik2020shape} or combinations of some of these \cite{muralikrishnan2019shape,poursaeed2020coupling}. 
All of the above representations have their pros and cons based on memory requirements and surface fitting precision.

We concentrated on one of the most popular representation method based on polygonal meshes. Mesh is a set of vertices joined together with edges that enable a piece-wise planar approximation of a surface. 

An object's mesh can be obtained with a transformation of a mesh on a unit sphere \cite{spurek2020hypernetwork,spurek2020hyperflow,wang2018pixel2mesh}. However, such methods are limited, and they reconstruct objects that are topologically the same as spheres.  

\paragraph{Patch-based representations}
Patch-based approaches \cite{yang2018foldingnet,groueix2018papier,bednarik2020shape,deng2020better} are much more flexible and enable modeling virtually any surfaces, including those with a non-disk topology. It is achieved using parametric mappings to transform 2D patches into a set of 3D shapes. The first deep neural network which uses 2D manifold into 3D space was FoldingNet~\cite{yang2018foldingnet}. FoldingNet uses a single patch to model the surface of an object.

In AtlasNet~\cite{groueix2018papier}, the authors introduced a method that uses several patches to model a mesh.  The authors train simultaneously $k$ functions $\phi_1, \ldots, \phi_k$ that jointly constitute an {\em atlas}.
Each function transforms a square $(0, 1) \times (0, 1)$ into a neighborhood of a point from the object's surface. Elements in the atlas are trained independently. Consequently, these maps are not stitched together, causing discontinuities appearing as holes or intersections patches.

To address the problem mentioned above, most of the methods extend the Chamfer loss function of basic AtlasNet with additional terms. \citet{bednarik2020shape} added terms to prevent patch collapse, reduce patch overlap and calculate the exact surface properties analytically rather than approximating them. \citet{deng2020better} introduced two additional terms to increase global consistency of the local mappings explicitly. One of them exploits the surface normals and requires that they remain locally consistent when estimated within and across the individual mappings. Another term enforces better spatial configuration of the mappings by minimizing a stitching error. 

Although these modifications improve the quality of obtained results, their objective is to fix the deformations after patches' stitching.
In this paper we propose a different approach to solve such a problem - we reformulate the classical definition of atlas to obtain maps which are correctly connected. Therefore, our method tries to suppress the issue before it even occurs in the first place.

\paragraph{Autoencoder-based generative model for 3D point clouds}

Let $\X = \{X_i\}_{i=1,\ldots,n} $  be a given data set containing point clouds. The basic aim of an autoencoder is to transport the data through a latent space $\Z \subseteq \R^D$ while minimizing the reconstruction error. Thus, we search for an encoder $\E:\X \to \Z$ and decoder $\D:\Z \to \X$, which minimize the reconstruction error between $X_i$ and its reconstructions~$\D(\E(X_i))$.

For the point cloud representation, the crucial step is to define reconstruction loss that can be used in the autoencoding framework. In the literature, two distance measures are successively applied: Earth Mover’s (Wasserstein) Distance \cite{rubner2000earth}, and Chamfer pseudo-distance \cite{tran20133d}.

In the autoencoder-based generative model, we additionally ensure that the data transported to the latent comes from a chosen prior distribution~\cite{kingma2013auto,tolstikhin2017wasserstein,knop2020cramer,zamorski2020adversarial}.

\paragraph{Hypernetwork}

Hypernetworks \cite{ha2016hypernetworks} are defined as neural models that generate weights for a separate target network solving a specific task. 
The authors aim to reduce the number of trainable parameters by designing a hypernetwork with fewer of parameters than the original network. Making an analogy between hypernetworks and generative models, \citet{sheikh2017stochastic} use that
mechanism to generate a diverse set of target networks approximating the same function. 

\section{Local surface parametrization}\label{sec:math}

In this section, we introduce Continuous Atlas, a novel paradigm for creating meshes from patches. We are pointing out the limitations of current approaches that are based on Discrete Atlas representations and show how we overcome them using our model.

\begin{figure}
\begin{center} 
 \includegraphics[height=6.0cm]{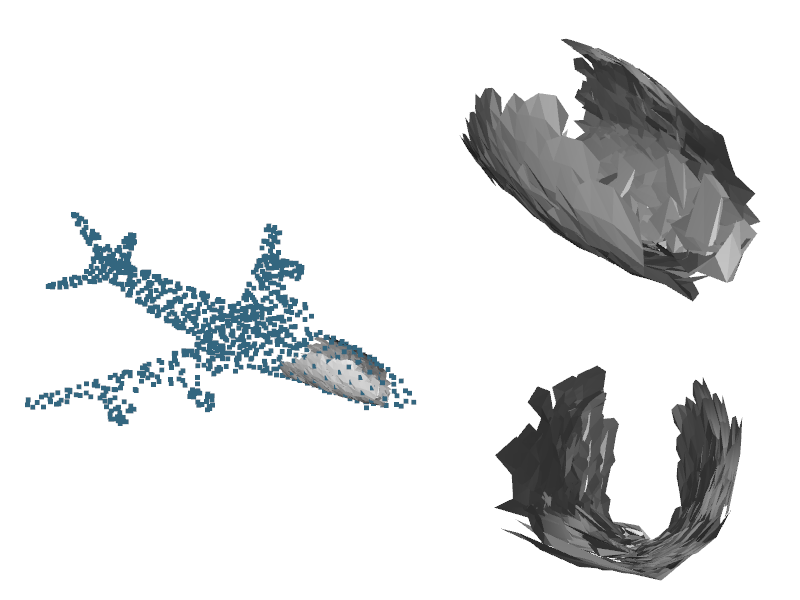}\\
 \includegraphics[height=3.0cm]{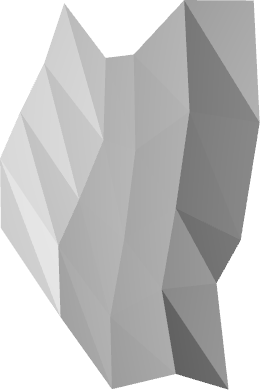}
 \includegraphics[height=3.0cm]{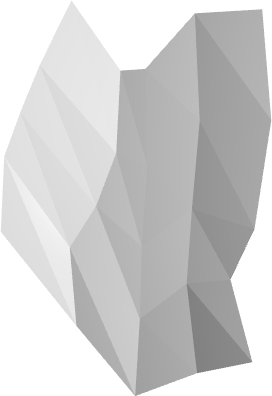}
 \includegraphics[height=3.0cm]{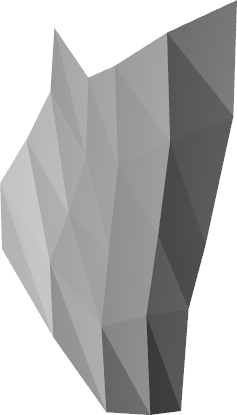}
 \includegraphics[height=3.0cm]{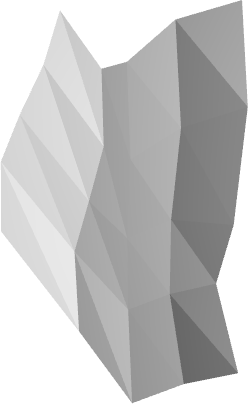}
\end{center} \vspace{-0.3cm}
\caption{Visualization of patches from the airplane. Parts next to each other are structurally similar and construct smooth surfaces.}
\vspace{-0.3cm}
\label{fig:part_mesh_a} 
\end{figure} 

\begin{figure*}[htbp]
\begin{center} 
  \includegraphics[width=0.9\linewidth]{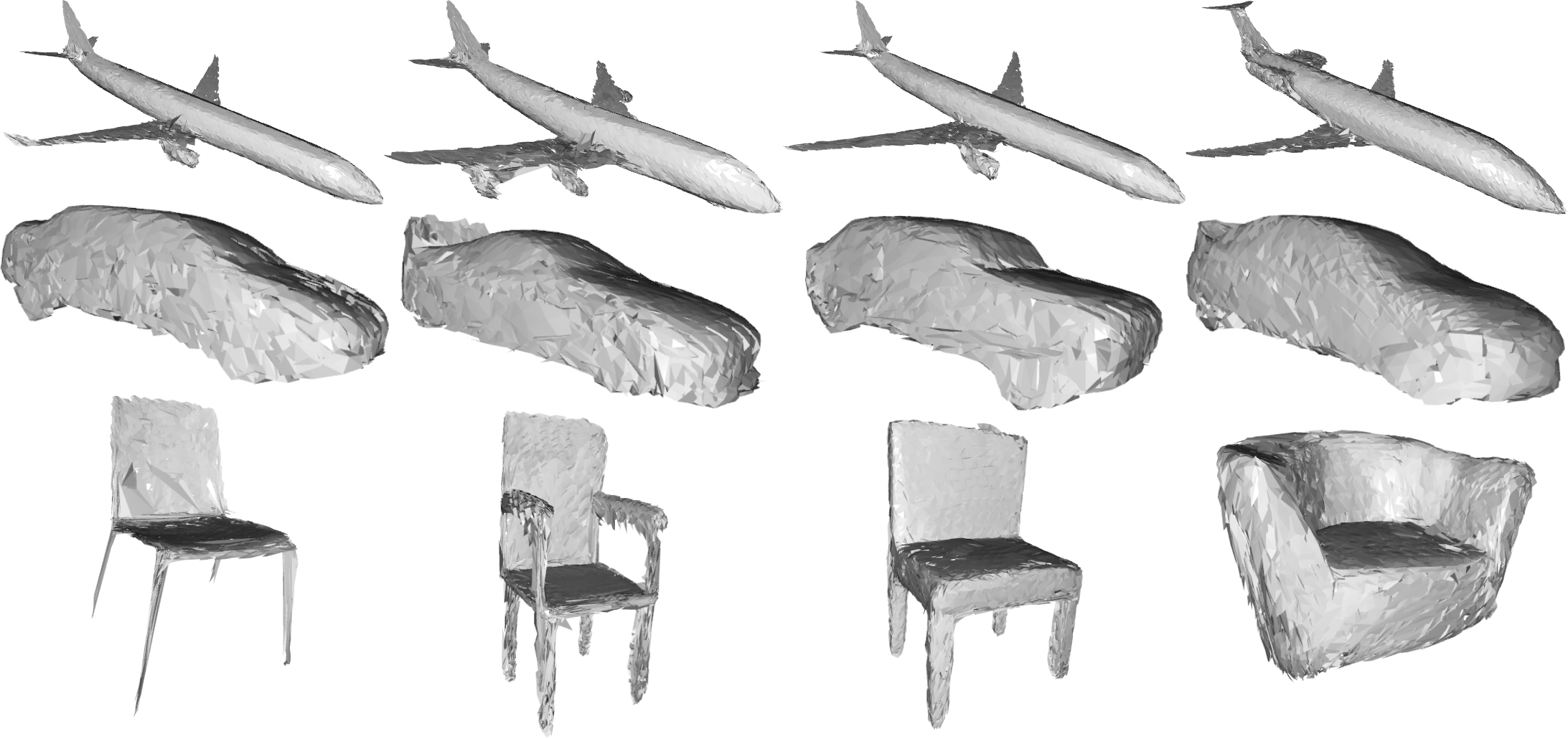}
\end{center} 
\caption{Mesh representations generated by our \our{} (HyperCloud) method.}

\vspace{-0.3cm}
\label{fig:rec_mesh} 
\end{figure*}

\subsection{Discrete Atlas}

We consider the definition of mesh introduced in \cite{groueix2018papier}. A 3D point cloud $S \subset \R^3$ is defined as a $2$-manifold (also called surface) if for every point $p \in S$ there is an open set $U$ in $\R^2$ and an open set $V$ in $\R^3$ containing $p$ such that $S \cap V$ is homeomorphic to $U$. The set homeomorphism from $S \cap V$ to $U$ is called a {\em chart}, and its inverse is a {\em parameterization}. A set of charts, such that their images cover the $2$-manifold, is called an {\em atlas} of the $2$-manifold.

Most of existing methods concentrated on direct modeling of {\em atlas} \cite{groueix2018papier,bednarik2020shape,deng2020better}. The authors train simultaneously $k$ functions $\phi_1, \ldots, \phi_k$ that jointly constitute a {\em discrete atlas}.
Each function $\phi_i$ transforms an open set $U \in \R^2$  into a neighborhood $V_i$ of a point $p_i \in S$ on the object's surface $S$. The neighborhood $V_i$ is a set of points around $p$ according to the Euclidean distance. In most of related works, $U$ is represented as a square $(0, 1) \times (0, 1)$.

Theoretically, such an approach should reconstruct a single smooth mesh. However, the model operates on an arbitrary given number of $k$ discrete functions, where a single function is responsible for generating a single patch. Consequently, it produces the discrete number of $k$ patches that are disjoint in the end.  It means that it creates empty places on an object's surface since patches can be disconnected. From a practical point of view, such a mesh is undesirable and requires further post-processing to be used in real-world applications. 

An atlas $\A(S)$ containing $k$ charts can be defined as a set of pairs 
$
\A(S) = (\phi_i, V_i)_{i=1,\ldots,k},
$
such that:

\begin{center}
$
\bigcup_{i=1}^k \phi_i( U ) = \bigcup_{i=1}^k V_i = S
$
\end{center}

where $\phi_i$ can be a Multilayer Perceptron. Practically, each chart $\phi_i(\cdot; \theta_i)$ is parametrized by learnable parameters $\theta_i$ and produces a part of the reconstructed object independently of others. All models $\phi(\cdot ;\theta_1), \ldots, \phi(\cdot; \theta_k)$ are learned together by using a global reconstruction cost function:

\begin{center}
$
Cost(S; \theta_1, \ldots, \theta_k) = \L \left(\bigcup\limits_{i=1}^k \phi( U; \theta_i ), S \right),
$
\end{center}

where $\L$ is either the Chamfer or the Earth-Mover distance.

\begin{figure*}[]
\centering
 \includegraphics[width=0.95\linewidth]{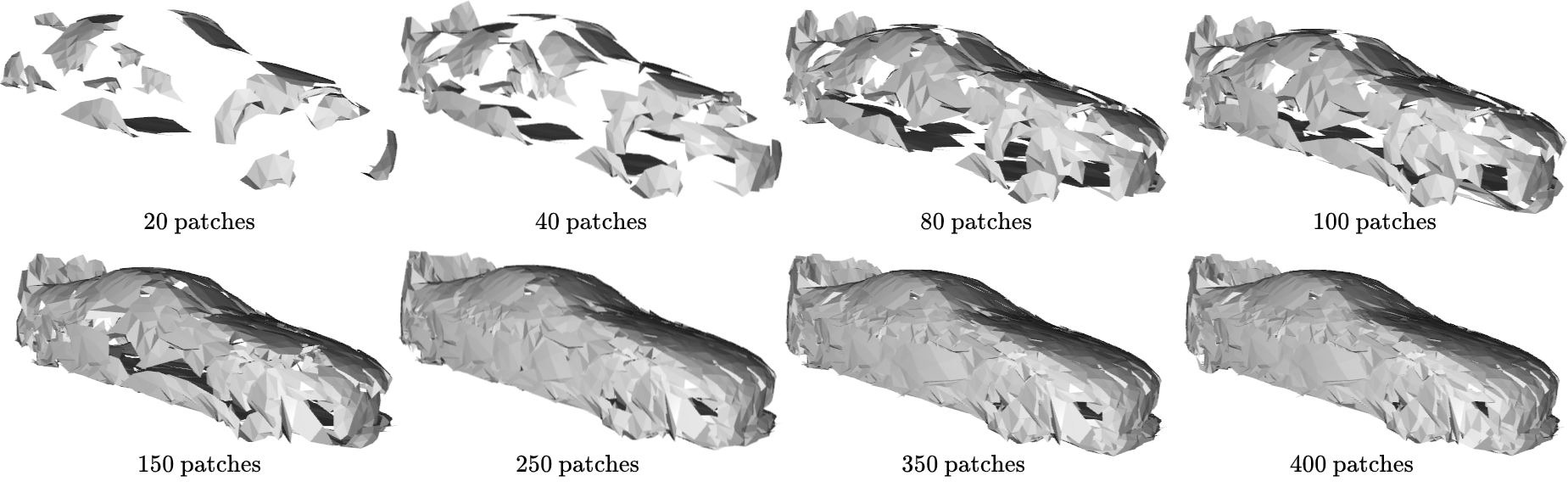}
\vspace{-0.3cm}
\caption{Mesh representations generated by our \our{} (HyperCloud) method with different number of patches.}
\vspace{-0.3cm}
\label{fig:patches_mesh} 
\end{figure*}

The product of all charts is further merged to obtain the reconstructed object.

Each map $V_i = \phi_i(U)$ models a neighbor of a point $p_i \in S$. Such an atlas of the object $S$ and a subset of points $P \subset S$ can be defined as a set:
$
\A(S, P) = (\phi_i, p_i)_{i=1,\ldots,k},
$
where $p_i \in P$, $\phi_i : U \to V(p_i) \subset S $, $V(p_i)$ is a neighborhood of $p_i$
and:  
$$
\bigcup_{i=1}^k \phi_i( U ) = \bigcup_{i=1}^k V(p_i) = S.
$$
Using this formulation, charts are trained to approximate the target surface as closely as possible. However, it does not consider the stitching process itself - no information is shared between patches. If one of them fails to cover the neighborhood of $p$ properly, then no other patch will fix that part.

\subsection{Continuous Atlas}
To mitigate the issue of the discrete atlas, we define \textit{Continuous Atlas}, a novel paradigm for meshing any object with an atlas that is leveraged in our method. In the first step, we construct a mapping that models a local structure of the object $S$. By {\em Continuous Atlas} ($\CA{}$), we define a mapping $\phi$ which transforms an open set $U \subset \R^2$ and a point $p \in S$ into a local neighborhood $V(p) \subset S$ of point $p$:
$$
\CA(S) = \{ \phi : (U, p) \to V(p) \subset S, \mbox{ for all } p \in S \}
$$

Instead of using set of discrete functions $\phi(\cdot ;\theta_1), \ldots, \phi(\cdot; \theta_k)$ we utilize only one transformation $\phi$ which locally models the surface of the object. We achieve that by providing an additional argument $p$ to the transformation function $\phi$ that is conditioning the positioning and shape of a 3D patch on the surface of an actual object it models. In contrast to a traditional conditioning mechanism in \mbox{AtlasNet}, $p$ is not a global descriptor of the object but a direct point of the desired surface $S$.

We extend this definition further to use any point $p_i \in S$ and produce patches in any place on the object. Therefore, we can choose any elements as points $P \subset S$ and produce an atlas containing pairs: 
\begin{equation}
    \label{eq:ca}
    \CA(S, P) = (\phi(\cdot, p_i), p_i)_{i=1,\ldots,k}.
\end{equation}

The proposed framework overcomes the limitations of previous methods. First, we theoretically solve the problem of stitching partial meshes since every chart is informed about its local neighborhood. Second, our method can easily fill the missing spaces in the final mesh by adding a new mapping for the region of interest. Because we can create an infinite number of patches using our approach, it is sufficient to locate a point in the empty space neighborhood and create an additional patch using $\phi$ function conditioned on the selected point.

We present in Fig.~\ref{fig:part_mesh_a} that our method can smoothly stitch patches on an example airplane object. 
Since the surface of the object is smooth, our map $\phi \in \CA(S)$ is also continuous with respect to the $p_i$ argument in Eq. (\ref{eq:ca}).

\section{\our{}: Local approximation of surface}
\label{sec:method}

We present Locally Conditioned Atlas (\our{}), a framework for generating and reconstructing meshes of objects using an atlas of localized charts that leverage the introduced notion of the continuous atlas. It consists of two parts. Firstly, we map the target object into a known prior distribution (training) or sample points from that prior (evaluation). This step is realized with a hypernetwork. Secondly, we use a separate neural network that transforms a point from that prior concatenated with points from a 2D square. Its goal is to place that patch on the object's surface being reconstructed.

\begin{figure*}[]
\begin{center} 
 \includegraphics[width=0.15\linewidth, trim=140 100 70 100, clip]{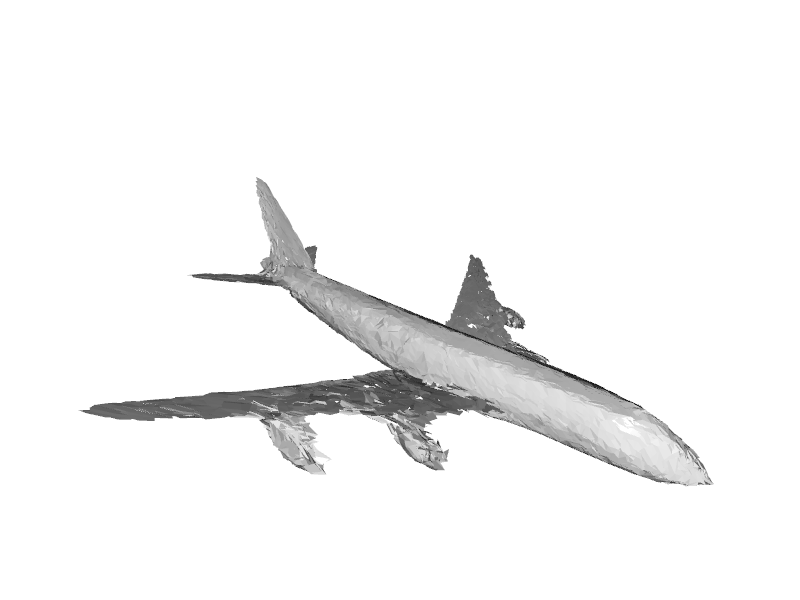}
 \includegraphics[width=0.15\linewidth, trim=140 100 70 100, clip]{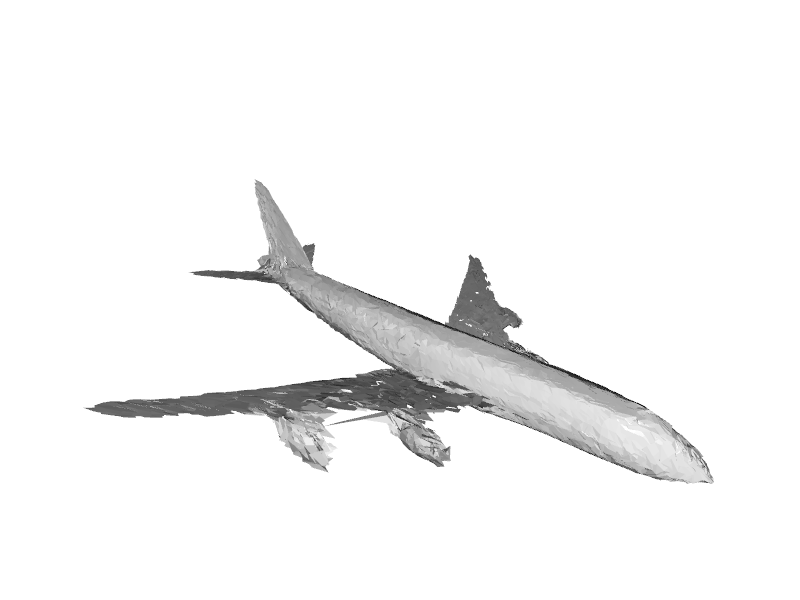}
 \includegraphics[width=0.15\linewidth, trim=140 100 70 100, clip]{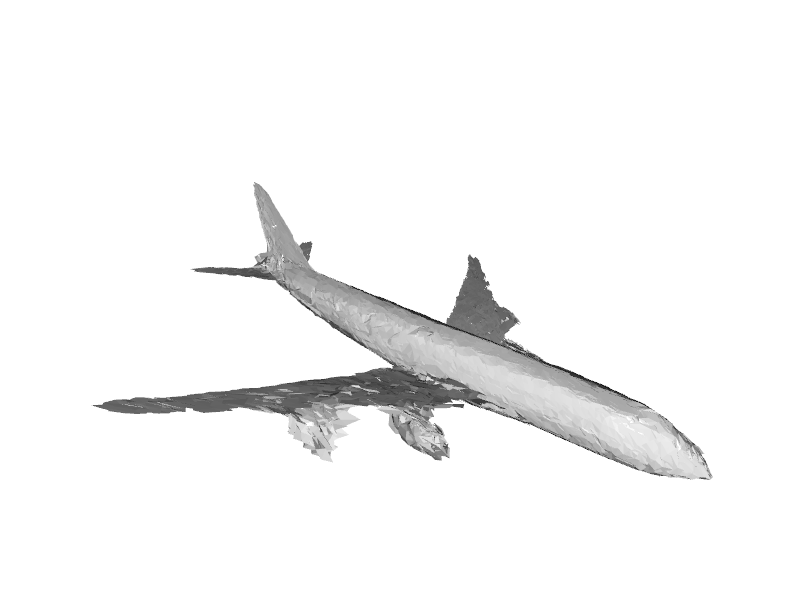}
 \includegraphics[width=0.15\linewidth, trim=140 100 70 100, clip]{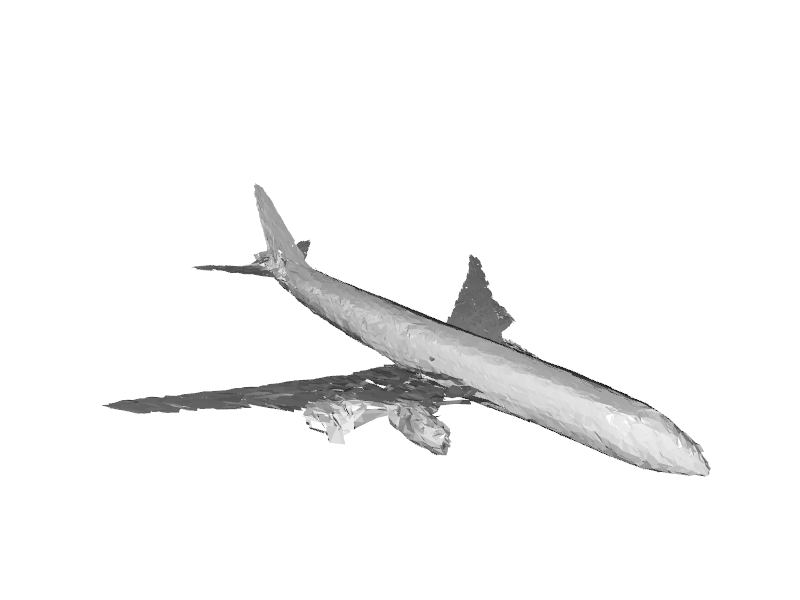}
 \includegraphics[width=0.15\linewidth, trim=140 100 70 100, clip]{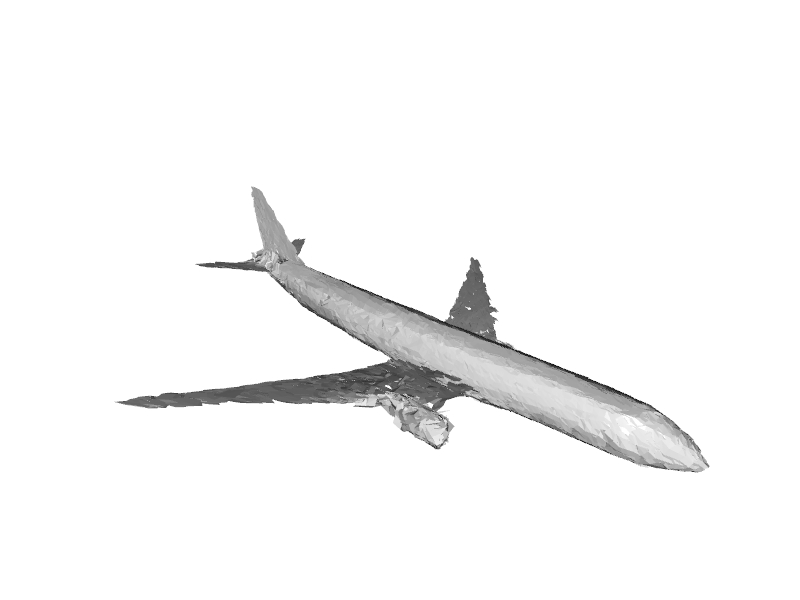}
  \includegraphics[width=0.15\linewidth, trim=140 100 70 100, clip]{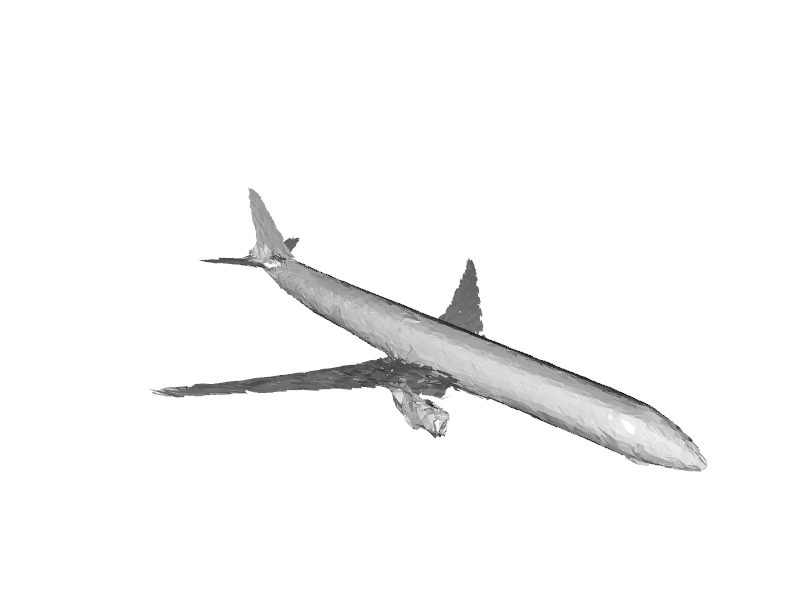}
 \\
 \includegraphics[width=0.15\linewidth, trim=10 10 10 10, clip]{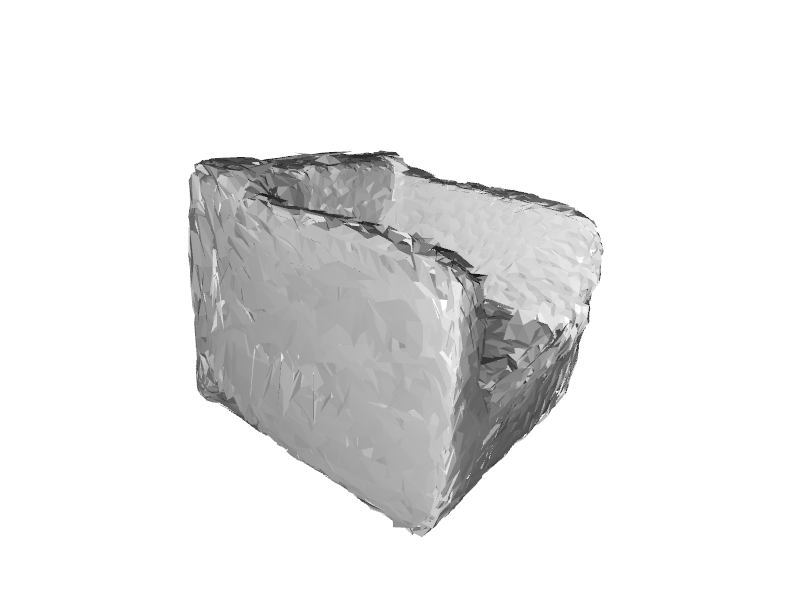}
 \includegraphics[width=0.15\linewidth, trim=10 10 10 10, clip]{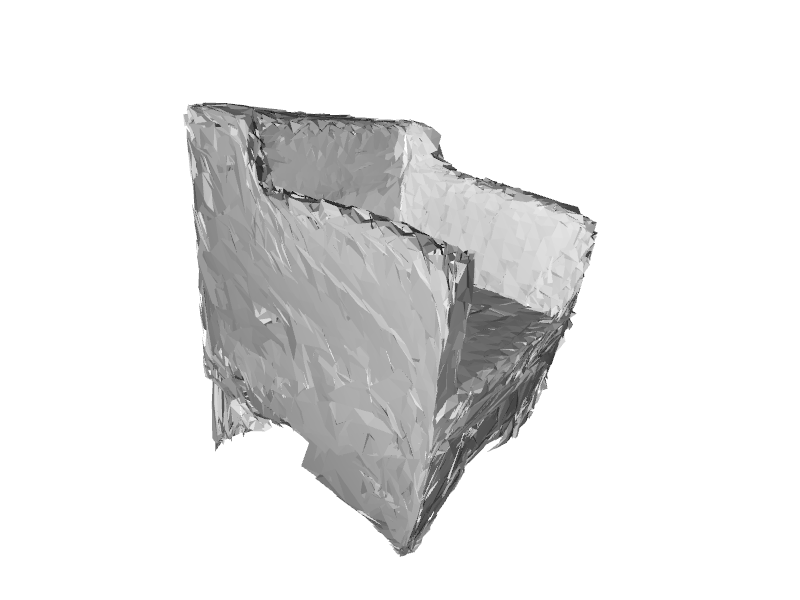}
 \includegraphics[width=0.15\linewidth, trim=10 10 10 10, clip]{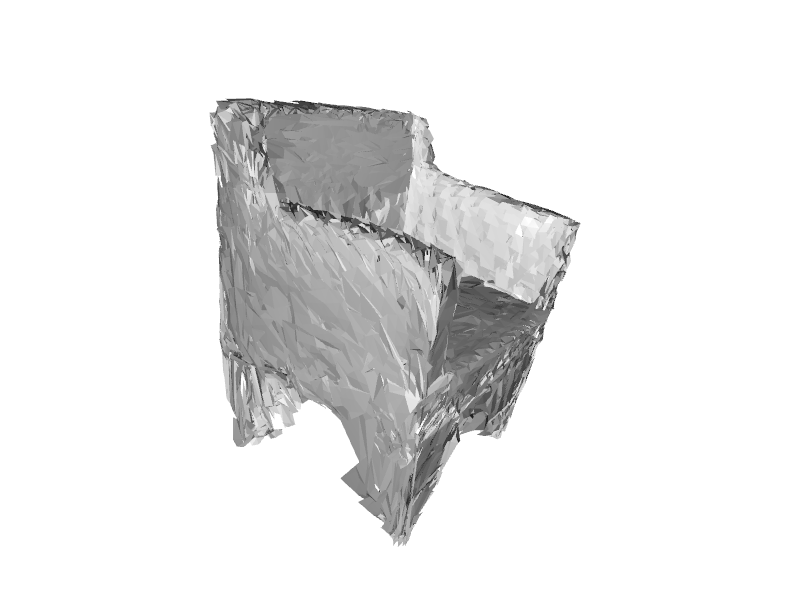}
 \includegraphics[width=0.15\linewidth, trim=10 10 10 10, clip]{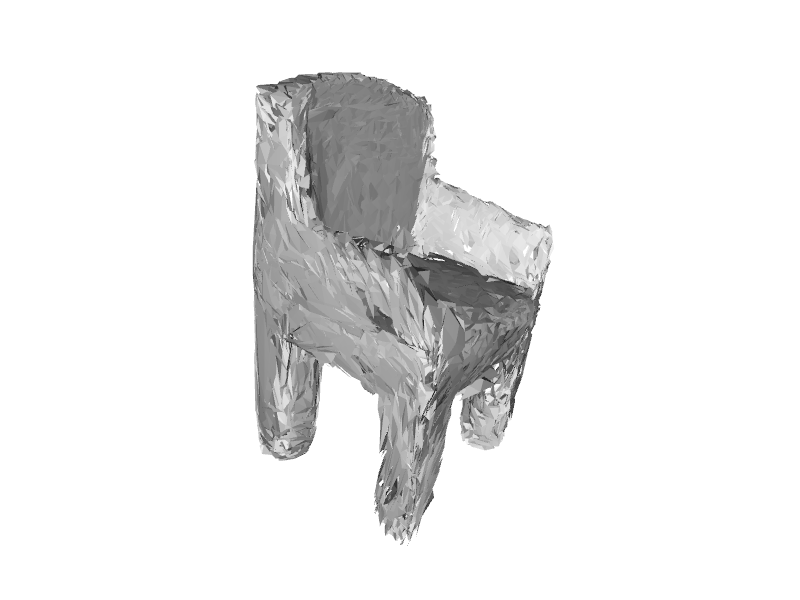}
 \includegraphics[width=0.15\linewidth, trim=10 10 10 10, clip]{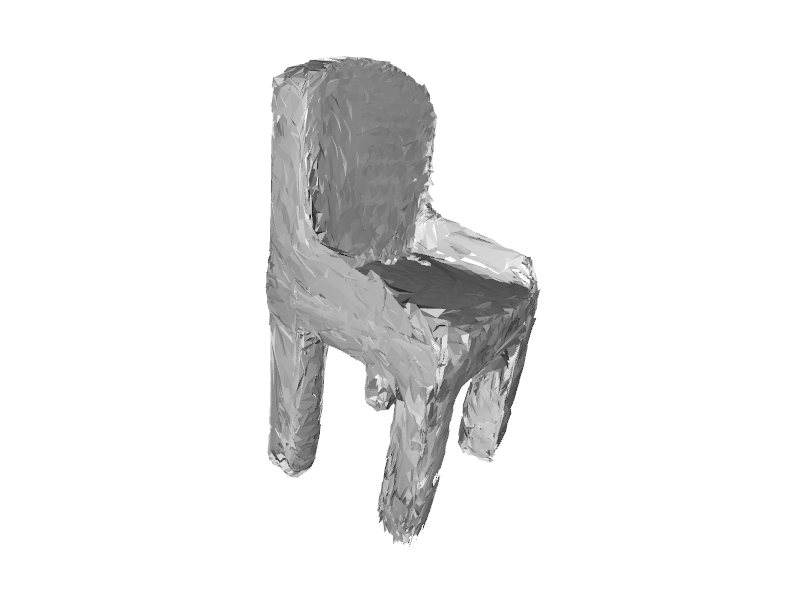}
  \includegraphics[width=0.15\linewidth, trim=10 10 10 10, clip]{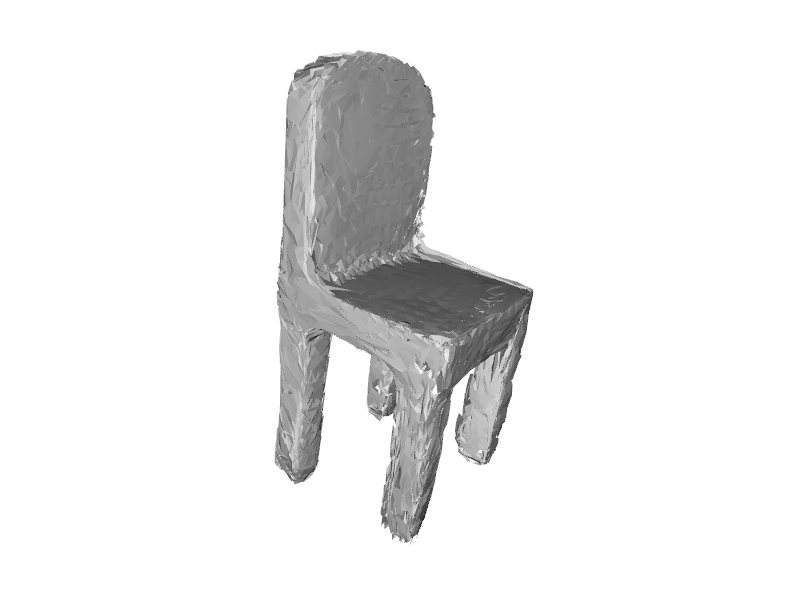} 
   \\
 \includegraphics[width=0.15\linewidth, trim=50 50 50 50, clip]{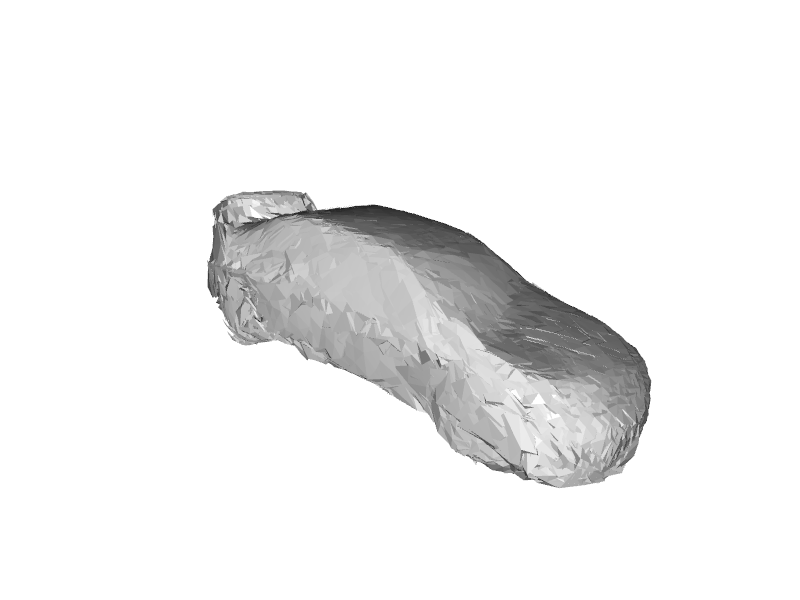}
 \includegraphics[width=0.15\linewidth, trim=50 50 50 50, clip]{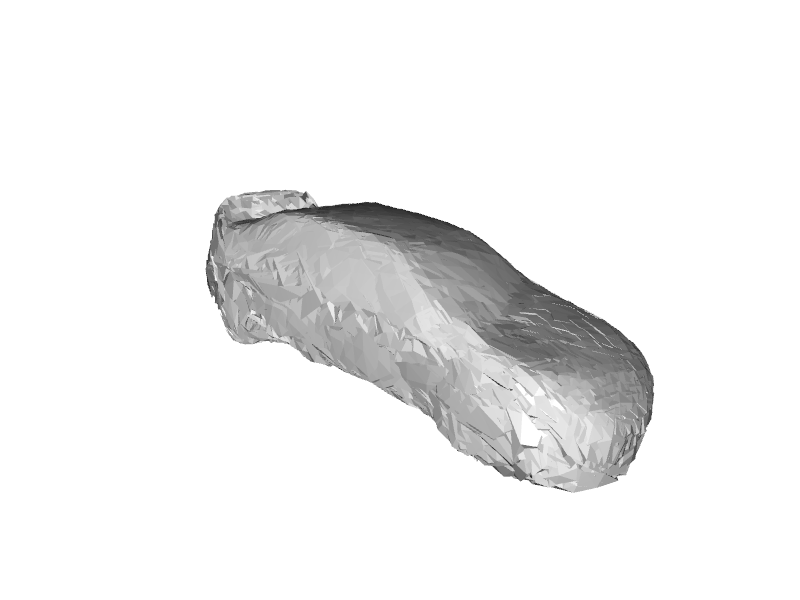}
 \includegraphics[width=0.15\linewidth, trim=50 50 50 50, clip]{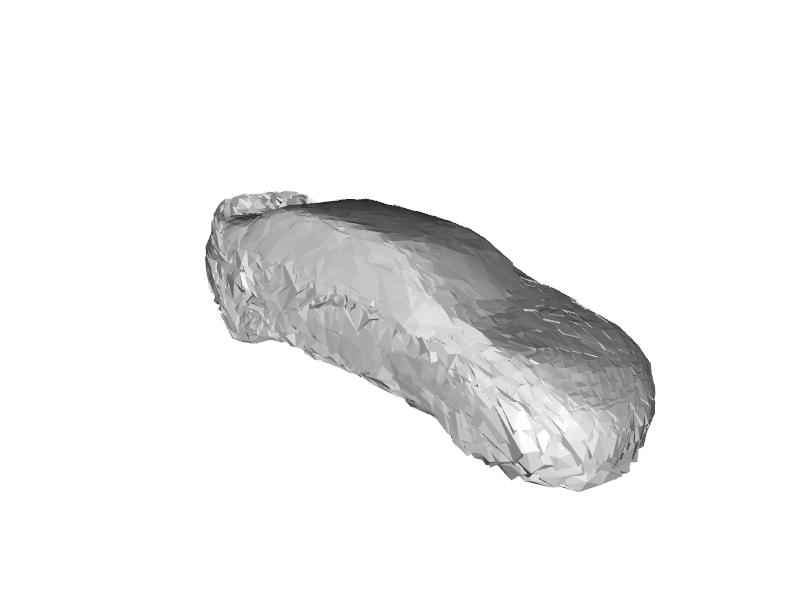}
 \includegraphics[width=0.15\linewidth, trim=50 50 50 50, clip]{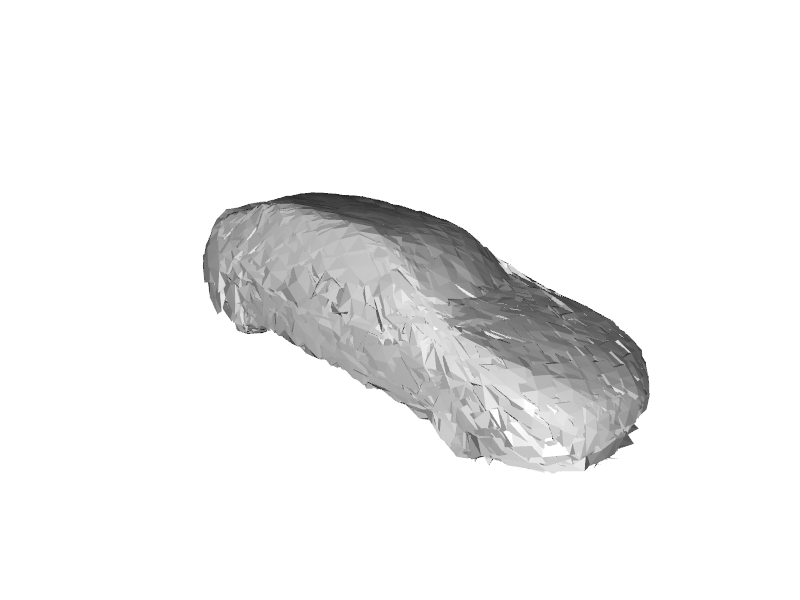}
 \includegraphics[width=0.15\linewidth, trim=50 50 50 50, clip]{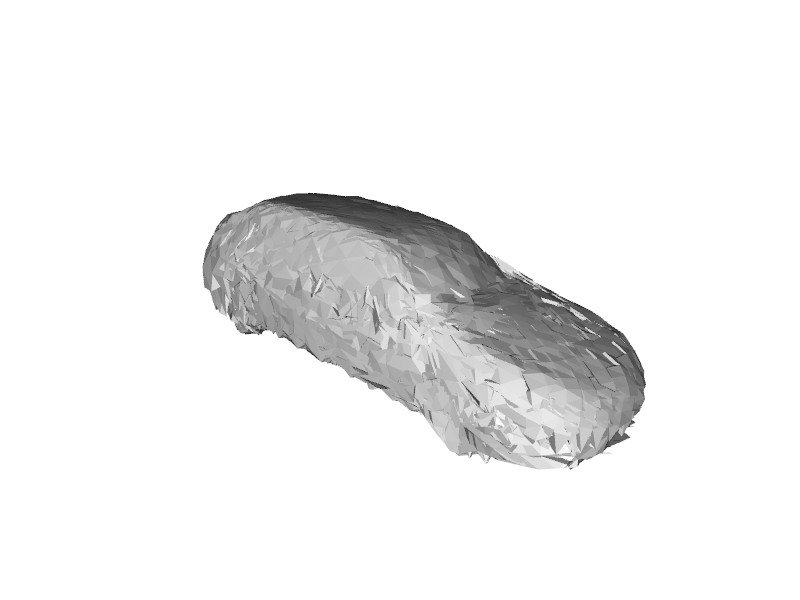}
  \includegraphics[width=0.15\linewidth, trim=50 50 50 50, clip]{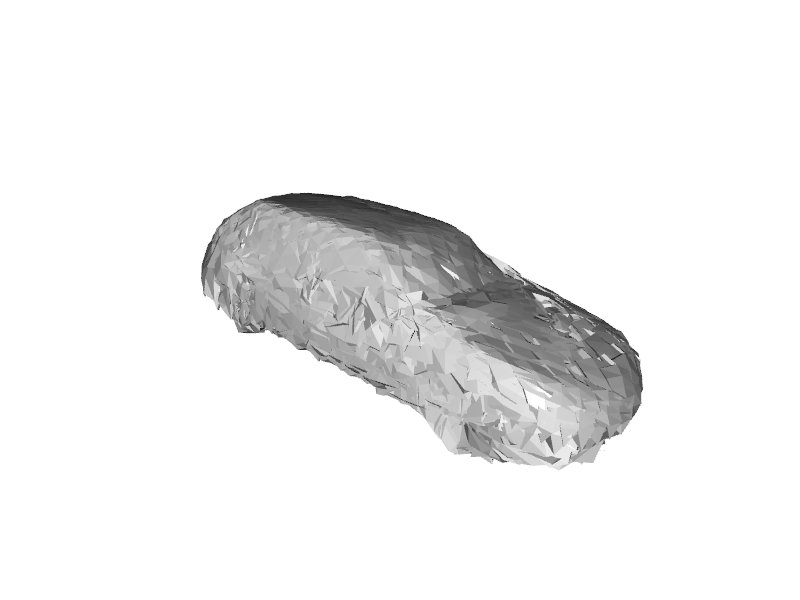} 
\end{center} 
\vspace{-0.3cm}
\caption{Mesh interpolation generated by our \our{} (HyperCloud) method.}
\vspace{-0.3cm}
\label{fig:inter_mesh} 
\end{figure*}

\paragraph{Part A: Generative auto-encoder using  hypernetworks} 
Since we directly operate on points lying on surfaces of 3D objects, we use an existing solution based on hypernetworks \mbox{HyperCloud}~\cite{spurek2020hypernetwork} or \mbox{HyperFlow}~\cite{spurek2020hyperflow}\footnote{We can also use conditioning framework introduced in \cite{yang2019pointflow,chen2020wavegrad}} instead of the classical encoder-decoder architecture. 
These architectures produce weights of small neural networks that map a prior distribution of points into 3D objects. Since they map points from a known prior distribution, we can sample implicitly any $p_i \in S$ without having access to the actual target object after the training.  In our framework, we assume that the hypernetwork (part A in Fig. \ref{fig:idea}) is already pretrained so it does not contribute to the total training time.

HyperCloud and HyperFlow use a hypernetwork to output weights of a generative network for 3D point clouds instead of directly generating these point clouds. More specifically, the surface of a 3D object is represented as a parametrized function $S: \R^3 \to \R^3$, which returns a point on the surface of the objects given a point~$(x, y, z)$ from the prior distribution.

In our framework, we use a hypernetwork
$
\begin{array}{c}
H: \mathbb{R}^3 \supset X \to W_T ,
\end{array}
$
which, for a point-cloud $X \subset \R^3$, returns weights $W_T$ to the corresponding target network $T$.
Thus, a point cloud $X$ is represented by a function 
$$
\begin{array}{c}
T((x,y,x);W_T) = T((x,y,x); H(X;W_H)).
\end{array}
$$
As a target network, we use a classical MLP in the case of HyperCloud, and Continuous Normalizing Flow (CNF) \cite{grathwohl2018ffjord} in HyperFlow.

To use the above model, we need to train the weights $W_H$ of the hypernetwork. In the case of HyperCloud, we minimize a loss between point clouds expressed as Chamfer Distance (CD) or Earth Mover’s Distance (EMD) over the training set of point clouds. More precisely, we take an input point cloud $X \subset \R^3$ and pass it to $H$. The hypernetwork returns weights $W_T$ to target network $T$, which reconstructs the object. Next, the input point cloud $X$ is compared with the output from the target network $T$ (see part A in Figure~\ref{fig:idea}). 
HyperFlow works similarly but uses log-likelihood as a cost function.

\paragraph{Part B: Locally Conditioned Atlas} 
\our{} implements the introduced paradigm of \textit{Continuous Atlas}. It consists of a $\phi$ function which transforms an open set $U \subset \R^2$ and a point $p \in S$ into a local neighborhood $V(p) \subset S$ of a point $p$:                
$$
\CA(S) = \{ \phi : (U , p) \to V(p) \subset S, \mbox{ where } p \in S \}
$$
where $U \subset \R^2$ is an open set $U = (0, 1) \times (0, 1)$. To model the transformation $\phi$ with a neural network, we have to simultaneously:
\begin{enumerate}[label=\protect\circled{\arabic*}]
\item transform a uniform distribution on $U = (0, 1) \times (0, 1) \subset \R^2$ to surface the object,
\item model a local neighborhood of an arbitrary element $p \in S$.
\end{enumerate}
Therefore, we use a hypernetwork that produces parameters of a small neural network that performs \circled{1}, and the conditioning of that neural network with a point $p$ to realize \circled{2}. The transformation $\phi$ is a fully connected network and is formulated as:
\begin{equation*}
\{\phi([u, p]; W_{\phi})| u \in U \} = V(p),
\end{equation*}
where $W_{\phi}$ are weights of $\phi$ produced by the hypernetwork directly from the point cloud embedding and $[\cdot, \cdot]$ is a concatenation operator.   

\begin{table*}
\centering
\caption{Generation results. MMD-CD scores are multiplied by
$10^3$; MMD-EMD and JSD scores are multiplied by $10^2$. (HC) denotes the HyperCloud autoencoder in \our{}, and (HF) - the HyperFlow autoencoder. For HyperCloud and HyperFlow, we use both variants of the models that generate point clouds (P) and meshes (M). }
\scalebox{0.8}{
\begin{tabular}{cccccccccccccccc}
\toprule
 \multirow{3}[4]{*}{Method}  & \multicolumn{5}{c}{\textit{Airplane}} & \multicolumn{5}{c}{\textit{Chair}} & \multicolumn{5}{c}{\textit{Car}} \\
\cmidrule(lr){2-6}
\cmidrule(lr){7-11}
\cmidrule(l){12-16}

& \multirow{2}[2]{*}{JSD} & \multicolumn{2}{c}{MMD} & \multicolumn{2}{c}{COV}
& \multirow{2}[2]{*}{JSD} & \multicolumn{2}{c}{MMD} & \multicolumn{2}{c}{COV}
& \multirow{2}[2]{*}{JSD} & \multicolumn{2}{c}{MMD} & \multicolumn{2}{c}{COV}\\  
\cmidrule(lr){3-4}
\cmidrule(lr){5-6}
\cmidrule(lr){8-9}
\cmidrule(lr){10-11}
\cmidrule(lr){13-14}
\cmidrule(l){15-16}
&  & CD & EMD & CD & EMD 
&  & CD & EMD & CD & EMD
&  & CD & EMD & CD & EMD\\ 
\midrule
\multicolumn{16}{c}{\emph{Point Cloud Generation}}\\
 \midrule
 l-GAN    & \bf 3.61 & 0.269 & 3.29 & \bf 47.90 &  50.62 & 2.27 & 2.61  &  7.85 & 40.79 & 41.69 & 2.21 & 1.48  & 5.43 & 39.20 & 39.77 \\
 PC-GAN  & 4.63 & 0.287 & 3.57 & 36.46 & 40.94 & 3.90 & 2.75  & 8.20 & 36.50 & 38.98 & 5.85 & 1.12  & 5.83 & 23.56 & 30.29\\
 PointFlow     & 4.92 & \bf 0.217 & 3.24 & 46.91 & 48.40 &  1.74 &  2.42  & 7.87 & \bf 46.83 & \bf 46.98 & \bf 0.87 & \bf 0.91  & 5.22 &  44.03 &  46.59\\
 HyperCloud(P) & 4.84 & 0.266 & 3.28 & 39.75 & 43.70 & 2.73 & 2.56  & \bf 7.84 & 41.54 & 46.67 & 3.09 & 1.07  & 5.38 & 40.05 & 40.05   \\ 
 HyperFlow(P) &  5.39    &   0.226     &  \bf 3.16   &    46.66   &    \bf 51.60 & \bf 1.50   &  2.30    &   8.01  & 44.71 &  46.37 &  1.07   &    1.14  &  5.30   & 45.74 &   47.44 \\
\midrule
  \multicolumn{16}{c}{\emph{Mesh Generation}} \\
\midrule
HyperCloud(M)  & 9.51  & 0.451 & 5.29 & 30.60    & 28.88  & 4.32  & 2.81 & 9.32 & 40.33 & 40.63   & 5.20  & 1.11 & 6.54 & 37.21 & 28.40  \\
HyperFlow(M)  & 6.55  &  0.384 &  3.65 & 40.49 &  48.64 &   4.26 &  3.33 &  8.27 &  41.99 & 45.32 & 5.77 & 1.39 &  5.91 & 28.40 &  37.21 \\ 
\our{}-HC &  16.1   & 0.664  & 4.71  &  30.37 & 32.59  & 4.45    &  3.03 & 8.55 & 42.45 & 38.22 & 1.91  & 1.13 & 5.50 & \bf 53.69 & \bf 50.56 \\
\our{}-HF &    4.80 &  0.223  &  3.20   &  44.69    &  47.91 & 2.54   &  \bf 2.23    &   7.94  & 43.35 &  42.60 & 1.16   & 0.92 & \bf 5.21 & 44.88 &   47.72 \\

\bottomrule
\end{tabular}
}
\label{tab:gen_results}
\vspace{-5.8pt}
\end{table*}

The transformation $\phi$ is modeled as a target network represented as MLP with weights $W_{\phi}$ produced by the hypernetwork $T_{\phi}$. Therefore, we can create an individual $\phi$ function for each of the 3D shapes and significantly reduce the number of parameters of the function by eliminating the need to share the parameters among the shapes. For $T_{\phi}$ we use the architecture analogical to $T$, but we train it with a different cost function. The new target network does not directly transfer uniform distribution on $U$ but uses conditioning as follows. 

Let $X_U = {(x_i,y_i)}_{i=1,\ldots,n}$ be a sample from a uniform distribution on $U = (0, 1) \times (0, 1) \subset \R^2$ and $ p = (p_x,p_y, p_z) \in X$. By $X_U(p)$ we define a set containing coordinates of a point $p$ concatenated with each element from $X_U$:
$$
X_U(p) = {(x_i,y_i, p_x,p_y, p_z)}_{i=1,\ldots,n} \subset \R^5.
$$

$\phi$ is a neural network that transfers $X_U(p)$ into a neighborhood of $p$:
$$
\phi: \R^5 \supset X_U(p) \to V(p).
$$

The main idea of mapping points to the neighborhood is to model a local manifold of data. One intuitive solution to determine neighbors is to use the K-nearest neighbors algorithm. The neighborhood of size $k$ of $p$ is defined as the $k$ closest elements of $p$ in $X$. Therefore we train \our{} in the second stage by using the following cost function:
$$
 \L \left( \phi( X_U(p) ; W_{\phi} ), V(p) \right),
$$
where $\L$ is a Chamfer or Earth-Mover distance and $p = T(x)$ is a randomly taken point from the reconstructed $X$, obtained in the first stage.

The above formulation alone causes that many of the produced patches have unnecessarily long edges, and the network folds them, so the patch fits the surface of an object. To mitigate the issue, we add an edge length regularization motivated by \cite{wang2018pixel2mesh}. If we assume that the reconstructed mesh has the form of a graph $M=(V, E)$ with edges $E$, then the term is defined as follows: 
$
l_{loc} = \sum_{e \in E}  ||e||^2_2,
$
where $||e||^2_2$ is a squared norm of length of an edge $e \in E$. The overall loss is a weighted sum of two parts:
$$
\L \left( \phi( X_U(p) ; W_{\phi} ), V(p) \right) + \lambda l_{loc},
$$
where $\lambda$ is a hyperparameter of the model.

\section{Experiments}

In this section, we describe the experimental results of the proposed method. First, we evaluate the generative capabilities of the model. Second, we provide the reconstruction result with respect to reference approaches. Finally, we check the quality of generated meshes, comparing our results to baseline methods. Throughout all experiments, we train models with Chamfer distance. We also set $\lambda = 0.0001$. We denote \our{}-HC when HyperCloud is used as the autoencoder architecture (Part A in Fig. \ref{fig:idea}) and \our{}-HF for the HyperFlow version.

\subsection{Generative capabilities}

\begin{table*}
\begin{center}
\caption{Shape auto-encoding on the ShapeNet dataset. The best results are highlighted in bold. CD is multiplied by $10^4$, and EMD is multiplied by $10^2$. (HC) denotes the HyperCloud autoencoder in \our{}, and (HF) - the HyperFlow autoencoder.}
\scalebox{0.8}{
\begin{tabular}{cccccccccc}
\toprule
\multirow{2}[2]{*}{Dataset} & \multirow{2}[2]{*}{Metric} & \multicolumn{2}{c}{l-GAN} & \multicolumn{2}{c}{ AtlasNet} & \multirow{2}[2]{*}{PointFlow} & \multirow{2}[2]{*}{\our{}-HC} &  \multirow{2}[2]{*}{\our{}-HF} &   \multirow{2}[2]{*}{\textcolor{gray}{Oracle}}  \\ 
\cmidrule(lr){3-4}
\cmidrule(lr){5-6}
                           &     & CD          & EMD       &  Sphere         & Patches    &  &    &  &                    \\ 
\midrule
\multirow{2}{*}{Airplane}                        
                           & CD & 1.020 & 1.196 & 1.002 & \bf 0.969 & 1.208 & 1.135 &  1.513  & \gray{0.837} \\
                           & EMD &  4.089 & \bf 2.577 & 2.672 & 2.612 & 2.757 & 2.881 & 2.990  & \gray{2.062} \\
                           \midrule
\multirow{2}{*}{Chair}                        
                           & CD & 9.279 & 11.21 & \bf 6.564 & 6.693 & 10.120 & 10.382 & 12.519  & \gray{3.201} \\
                           & EMD & 8.235 & 6.053 & 5.790 & \bf 5.509 & 6.434 & 6.738 &  6.973  & \gray{3.297}\\
                           \midrule
\multirow{2}{*}{Car }                        
                           & CD & 5.802 & 6.486 & \bf 5.392 & 5.441 & 6.531 & 6.575 & 7.247  & \gray{3.904}\\
                           & EMD & 5.790 & 4.780 & 4.587 & \bf 4.570 & 5.138 & 5.126 &   5.275 & \gray{3.251}\\
\bottomrule
\end{tabular}
}
\label{tab:rec_results}
\vspace{-11.8pt}
\end{center}
\end{table*}

We examine the generative capabilities of the provided \our{} model compared to the existing reference approaches. In this experiment, we follow the evaluation protocol provided in \cite{yang2019pointflow}.  We use standard measures for this task like Jensen-Shannon Divergence (JSD), coverage (COV), and minimum matching distance (MMD), where the last two measures are calculated for Chamfer (CD) and Earth-Mover (EMD) distances separately. 

We compare the results with the existing solutions that aim at point cloud generation: latent-GAN \cite{achlioptas2017learning}, PC-GAN \cite{li2018point}, PointFlow \cite{yang2019pointflow}, HyperCloud(P) \cite{spurek2020hypernetwork} and HyperFlow(P) \cite{spurek2020hyperflow}. We also consider in the experiment two baselines,  HyperCloud(M) and HyperFlow(M) variants, that are capable of generating the meshes from the unit sphere. 
We train each model using point clouds from one of the three categories in the ShapeNet dataset: \emph{airplane}, \emph{chair}, and \emph{car}. 

The results are presented in Table~\ref{tab:gen_results}. \our{}-HF obtains comparable results to the reference methods dedicated for the point cloud generation. It can be observed that values of evaluated measures for HyperFlow(P) and \our{}-HF (uses HyperFlow(P) as a base model in the first part of the training) are on the same level. It means that incorporating an additional step (part B.) dedicated to mesh generation does not negatively influence our model's generative capabilities. On the other hand, if we use HyperFlow to produce meshes directly using the procedure described in \cite{spurek2020hyperflow} (see results for HyperFlow(M)), the generative capabilities are significantly worse for considered evaluation metrics.

\subsection{Reconstruction capabilities}

In this section, we evaluate how well our model can learn the underlying distribution of points by asking it to autoencode a point cloud. We conduct the autoencoding task for 3D point clouds from three categories in ShapeNet (\textit{airplane}, \textit{car}, \textit{chair}). In this experiment, we compare \our{} with the current state-of-the-art AtlasNet \cite{groueix2018papier} where the prior shape is either a sphere or a set of patches. Furthermore, we also compare with l-GAN \cite{achlioptas2018learning} and PointFlow \cite{yang2019pointflow}. We follow the experiment set-up in PointFlow and report performance in both CD and EMD in Table~\ref{tab:rec_results}. Since these two metrics depend on the point clouds' scale, we also report the upper bound in the "oracle" column. The upper bound is produced by computing the error between two different point clouds with the same number of points sampled from the same ground truth meshes. It can be observed that \our{}-HC achieves competitive results with respect to reference solutions. All reference methods were trained in an autoencoding framework (non-generative variants), while both of \our{} are preserving generative capabilities in the experiment.

We additionally show in Fig.~\ref{fig:inter_mesh} reconstruction results obtained by decoding linearly interpolated latent vectors of two objects from each class. Our model generates coherent and semantically plausible objects for all interpolation steps.

\begin{table}[htbp]
    \centering
    \caption{Comparison between \our{} and the related AtlasNet approach in terms of watertightness (WT). Note that for a spherical prior, meshes are always watertight.}
    \scalebox{0.85}{
        \begin{tabular}{ccccc}
            \toprule
                Method & \textit{Airplane} &\textit{Chair} & \textit{Car} & \textit{Average} \\
            \midrule
                AtlasNet (25 patches) &  0.516 & 0.507  & 0.499  & 0.507 \\
                \our{}-HC  & \textbf{1.00} &  \textbf{1.00} &  \textbf{1.00} & \textbf{1.00} \\
                \our{}-HF  &  \textbf{1.00}  & \textbf{1.00}  & \textbf{1.00}  & \textbf{1.00} \\
            \bottomrule
        \end{tabular}
    }
    \label{tab:watertightness}
\vspace{-1.8pt}
\end{table}

\subsection{Mesh quality evaluation}
Finally, we empirically show the proposed framework produces high-fidelity and watertight meshes. It means that it solves the initial problem of disjoint patches occurring in the original AtlasNet \cite{groueix2018papier}. To evaluate the continuity of output surfaces, we propose to use the following metric.

\textbf{Watertigthness}  Typically, a mesh is referred to as being either watertight or not watertight. Since it is a true or false statement, there is no well-established measure to define the degree of discontinuities in the object's surface. To fill this gap, we propose a metric based on a simple, approximate check of whether a mesh is watertight - the parity test. The test says that any ray cast from infinity towards the object has to enter and leave the object. It is realized as checking whether the number of rays' crossings with all triangles in the mesh is an even number. If so, the ray is said to pass the parity test. The mesh is watertight if all rays pass the test.

To leverage that knowledge, we express watertigthness as a ratio of rays that passed the parity test to the total number of all casted rays. Firstly, we sample $N$ points $p \in \hat{S}$ from all triangles of the reconstructed object $\hat{S}$. Since each point is associated with a triangle it was sampled from, we use a corresponding normal $\hat{n}$ of its triangle and negate it to obtain a direction of a ray $R(\hat{S}) \ni r = -\hat{n}p$ towards the object. Then, we calculate the number of crossings $c(r)$ with all triangles. For each ray, we set 1 if it passes a test and 0 otherwise. We sum test results over all rays and divide by the number of rays to obtain the watertightness ($WT$) measure, which we formulate as:

\begin{center}
$
    WT(\hat{S}) = \frac{\sum_{r \in R(\hat{S})} \mathbb{I}[c(r) \bmod 2 = 0]}{|R(\hat{S})|}
$
\end{center}

In this experiment, we set $N = 10^5$. Using more rays had a negligible effect on the output value of $WT$ but significantly slowed the computation. We compared AtlasNet with \our{} applied to HyperCloud (HC) and HyperFlow (HF). We show the obtained results in Table \ref{tab:watertightness}. Note that AtlasNet cannot produce watertight meshes for any of the classes, limiting its applicability. On the other hand, \our{} creates meshes where all sampled rays pass the test.

\section{Conclusions}

In this paper, we introduced a novel approach for generating high-quality 3D meshes composed of 2D patches directly from raw point clouds. We presented a \textit{Continuous Atlas} paradigm that allows our model, Locally Conditioned Atlas, to produce an arbitrary number of patches to form a watertight mesh. The empirical evaluation of \our{} on three extensive experiments confirms the validity of our approach and its competitive performance.


\end{document}